\documentclass[letterpaper]{article} 
\usepackage{aaai23}  
\usepackage{times}  
\usepackage{helvet}  
\usepackage{courier}  
\usepackage[hyphens]{url}  
\usepackage{graphicx} 
\usepackage{natbib}  
\usepackage{caption} 
\usepackage{algorithm}
\usepackage{algorithmic}
\usepackage{makecell}
\usepackage{newfloat}
\usepackage{listings}
\DeclareCaptionStyle{ruled}{labelfont=normalfont,labelsep=colon,strut=off} 
\floatstyle{ruled}
\newfloat{listing}{tb}{lst}{}
\floatname{listing}{Listing}
\nocopyright 

\title{Mx2M: Masked Cross-Modality Modeling in Domain Adaptation for 3D Semantic Segmentation}
\author{
    Boxiang Zhang\textsuperscript{\rm 1}\textsuperscript{\rm 3}\thanks{Work done during an internship at Tencent.}\equalcontrib,
    Zunran Wang\textsuperscript{\rm 2}\equalcontrib\thanks{Corresponding authors.},
    Yonggen Ling\textsuperscript{\rm 2},
    Yuanyuan Guan\textsuperscript{\rm 1}\textsuperscript{\rm 3},\\
    Shenghao Zhang\textsuperscript{\rm 2},
    Wenhui Li\textsuperscript{\rm 1}\textsuperscript{\rm 3}\footnotemark[3]
}
\affiliations{
    \textsuperscript{\rm 1} College of Computer Science and Technology, Jilin University, Changchun, China\\
    \textsuperscript{\rm 2} Robotics X, Tencent, Shenzhen, China\\
    \textsuperscript{\rm 3} Key Laboratory of Symbolic Computation and Knowledge Engineer, Jilin University, Changchun, China\\
    zhangbx2113@mails.jlu.edu.cn, zran.wang@outlook.com, liwh@jlu.edu.cn
}

\usepackage{bibentry}
\usepackage{diagbox}
\usepackage{color}
\usepackage{xcolor}
\usepackage{multirow}
\usepackage{amssymb}
\usepackage{booktabs}
\usepackage{subcaption}

\begin{document}
\maketitle
\begin{abstract}
Existing methods of cross-modal domain adaptation for 3D semantic segmentation predict results only via 2D-3D complementarity that is obtained by cross-modal feature matching. However, as lacking supervision in the target domain, the complementarity is not always reliable. The results are not ideal when the domain gap is large. To solve the problem of lacking supervision, we introduce masked modeling into this task and propose a method \textbf{Mx2M}, which utilizes \textbf{m}asked \textbf{cross}-\textbf{m}odality \textbf{m}odeling to reduce the large domain gap. Our Mx2M contains two components. One is the core solution, cross-modal removal and prediction (xMRP), which makes the Mx2M adapt to various scenarios and provides cross-modal self-supervision. The other is a new way of cross-modal feature matching, the dynamic cross-modal filter (DxMF) that ensures the whole method dynamically uses more suitable 2D-3D complementarity. Evaluation of the Mx2M on three DA scenarios, including Day/Night, USA/Singapore, and A2D2/SemanticKITTI, brings large improvements over previous methods on many metrics.
\end{abstract}

\section{Introduction}
\begin{figure*}[t]
  \centering
  \includegraphics[width=0.85\linewidth]{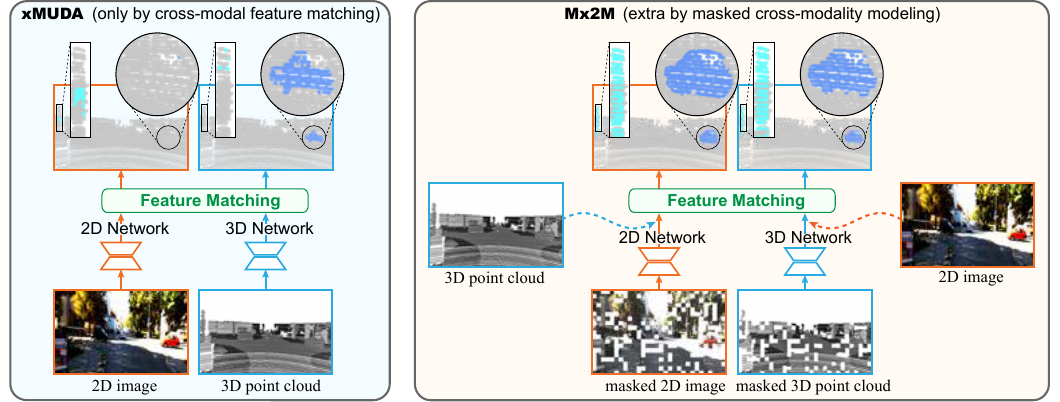}
  \caption{\label{fig:comparison} Left: the pipeline of xMUDA and its heirs (the segmentation results are from xMUDA). Right: the pipeline of our Mx2M. Thanks to masked cross-modality modeling, our method reduces the large domain gap and then achieves better segmentation results with the same backbone.}
\end{figure*}

3D semantic segmentation methods \citep{graham20183d,wang2019graph,hu2021bidirectional} often encounter the problem of shift or gap between different but related domains (\textit{e.g.} day and night). The task of cross-modal domain adaptation (DA) for 3D segmentation \citep{jaritz2020xmuda} is designed to address the problem, which is inspired by 3D datasets usually containing 2D and 3D modalities. Like most DA tasks, labels here are only available in the source domain, whereas the target domain has no segmentation labels. Existing methods, \textit{i.e.} xMUDA \citep{jaritz2020xmuda} and its heirs \citep{liu2021adversarial,peng2021sparse}, extract 2D and 3D features through two networks and exploit the cross-modal complementarity by feature matching to predict results. However, as lacking supervision in the target domain, the robustness of this complementarity is not good. As shown in the left part of Fig.\ref{fig:comparison}, if the domain gap is large and both networks underperform on the target domain, these methods appear weak.

The problem of lacking supervision once constricted the visual pre-training task and has been solved by methods with masked modeling \citep{he2021masked,bao2021beit,yu2021pointbert}, which has been proved to belong to data augmentation \cite{xu2022masked}. Its core solution is simple: removing a portion of inputs and learning to predict the removed contents. Models are fitted with sufficient data in this way, so that learn more inner semantic correspondences and realize self-supervision \citep{he2021masked}. For this DA task, this way of data augmentation and then the self-supervision can enrich the robustness and reduce the gap. Hence the idea is natural: if we introduce masked modeling into the task, the lacking supervision on the target domain and then the large gap are solved. Nevertheless, two problems are the key to introducing masked modeling. a) The core solution ought to be re-designed to fit for this task, where there are two modalities. b) For the cross-modal feature matching, we should explore a new way to suit the joining of masked modeling.

Given these observations, we propose a new method \textbf{Mx2M} utilizing \textbf{m}asked \textbf{cross}-\textbf{m}odality \textbf{m}odeling to solve the problem of lacking supervision for the DA of 3D segmentation. Our Mx2M can reduce the large domain gap by adding two new components to the common backbone for this task, which correspond to the above two problems. For the first one, we design the core solution in the Mx2M, cross-modal removal and prediction (xMRP). As the name implies, we inherit the 'removal-and-prediction' proceeding in the core solution of masked single-modality modeling and improve it with the cross-modal working manner for this task. During removal, the xMRP has two changes. i) Our CNN backbone cannot perform well with highly destroyed object shapes \citep{geirhos2020beyond}, so the masked portion is less. ii) To guarantee the existence of full semantics in this segmentation task, we do not mask all inputs and ensure at least one modality complete in each input. We can obtain the different xMRP by controlling the removal proceeding, which makes the Mx2M adapt to various DA scenarios. During prediction, to learn more 2D-3D correspondences beneficial to networks \citep{jaritz2020xmuda}, we mask images/points and predict the full content in points/images by two new branches. In this way, cross-modal self-supervision can be provided for the whole method. 

As for the second problem, we propose the dynamic cross-modal filter (DxMF) to dynamically construct the cross-modal feature matching by locations, which is inspired by impressive gains when dynamically establishing kernel-feature correspondences in SOLO V2 \citep{wang2020solov2}. Similarly, in our DxMF, we structure the 2D-3D kernel-feature correspondences. Kernels for one modality are generated by features from the other, which then act on features for this modality and generate the segmentation results by locations. With the joining of the DxMF, the Mx2M can dynamically exploit the complementarity between modalities. As is shown in the right part of Fig.\ref{fig:comparison}, with these two components, our Mx2M gains good results even in the scenario with a large domain gap. 

To verify the performance of the proposed Mx2M, we test it on three DA scenarios in \citep{jaritz2020xmuda}, including USA/Singapore, Day/Night, and A2D2/SemanticKITTI. Our Mx2M attains better results compared with most state-of-the-art methods, which indicates its effectiveness. In summary, our main contributions are as follows:
\begin{itemize}
\item We innovatively propose a new method Mx2M, which utilizes masked cross-modality modeling to reduce the large domain gap for DA of 3D segmentation. To our knowledge, it is the first time that masked modeling is introduced into a cross-modal DA task.
\item Two components are specially designed for this task, including xMRP and DxMF, which ensures the Mx2M effectively works and deals with various scenarios.
\item We achieve high-quality results on three real-to-real DA scenarios, which makes the Mx2M the new state-of-the-art method. The good results demonstrate its practicality.
\end{itemize}

\section{Related Work}

\paragraph{Domain Adaptation for 3D Segmentation.}Most works pay attention to DA for 2D segmentation \citep{zhang2021dobnet,zhang2020mfenet,li2019bidirectional}, which are hard to be applied to unstructured and unordered 3D point clouds. The DA methods for 3D segmentation \citep{qin2019pointdan,luo2020unsupervised,morerio2018minimal} are relatively few, but they also do not fully use the datasets that often contain both images and points. Hence, xMUDA \cite{jaritz2020xmuda} and its heirs \cite{liu2021adversarial,peng2021sparse} with cross-modal networks are proposed, which achieve better adaptation. Our Mx2M also adopts cross-modal networks, which has the same backbone as xMUDA.

\begin{figure*}[t]
  \centering
  \includegraphics[width=0.95\linewidth]{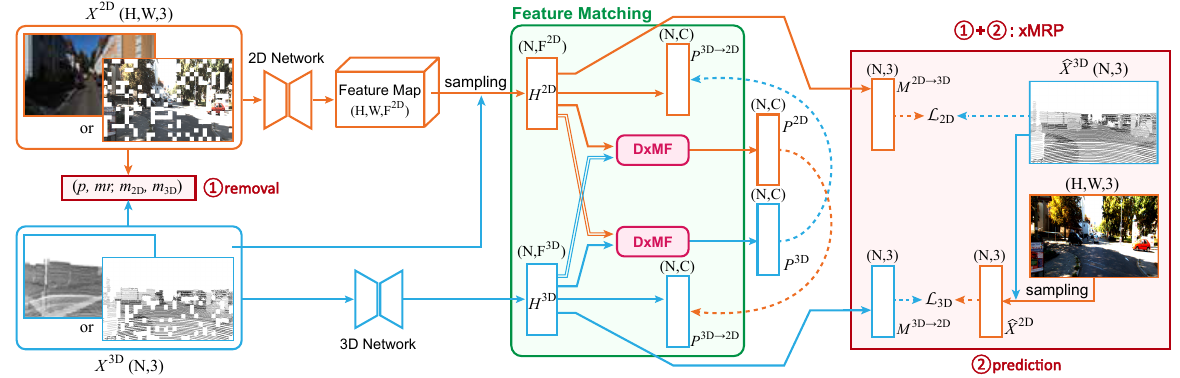}
  \caption{\label{fig:arch}The architecture of the Mx2M. We introduce masked modeling into the task of DA in 3D segmentation mainly by xMRP and DxMF. The former is the core solution where we remove (mask) \textcolor{orange}{images}/\textcolor{cyan}{points} and predict the full content in \textcolor{cyan}{points}/\textcolor{orange}{images}. The latter is for 2D-3D feature matching.}
\end{figure*}

\paragraph{Masked Modeling.}The masked modeling was first applied as masked language modeling \citep{kenton2019bert}, which essentially belongs to data augmentation \cite{xu2022masked}. Nowadays, it has been the core operation in self-supervised learning for many modalities, such as masked image modeling \citep{bao2021beit,xie2021simmim}, masked point modeling \citep{yu2021pointbert}, and masked speech modeling \citep{baevski2020wav2vec}. Their solutions are the same: removing a portion of the data and learning to predict the removed content. The models are fitted with sufficient data in this way so that the lacking of supervision is satisfied. Our Mx2M designs the masked cross-modality modeling for DA in 3D segmentation that uses point and image.

\paragraph{Cross-modal Learning.} Cross-modal learning aims at taking advantage of data from multiple modalities. For visual tasks, the most common scene using it is learning the 3D task from images and point clouds \citep{jing2021self,hu2021bidirectional,genova2021learning,liu2021learning,xu2021image2point,dai20183dmv,liu20213d}. The detailed learning means are various, including 2D-3D feature matching \citep{jing2021self,dai20183dmv,liu20213d,xu2021image2point}, 2D-3D feature fusion \citep{hu2021bidirectional}, 2D-3D cross-modal supervision \citep{genova2021learning,liu2021learning}, \textit{etc.} Besides, there are also some works conducting cross-modal learning on other modalities, such as video and medical image \citep{carreira2017quo,shan20183}, image and language \citep{lu2021pretrained,radford2021learning,fu2021violet}, as well as video and speech \citep{gao2021visualvoice,lee2021looking,wu2022time}. Cross-modal learning is also exploited in our M2xM: the core procedure xMRP leverages the cross-modal supervision, while the DxMF works in the way of 2D-3D feature matching.

\section{Method}
Our Mx2M is designed for DA in 3D segmentation assuming the presence of 2D images and 3D point clouds, which is the same as xMUDA \citep{jaritz2020xmuda}. For each DA scenario, we define a source dataset $\mathcal{S}$, each sample of which contains a 2D image $X^{\rm{2D,S}}$, a 3D point cloud $X^{\rm{3D,S}}$, and a corresponding 3D segmentation label $Y^{\rm{3D,S}}$. There also exists a target dataset $\mathcal{T}$ lacking annotations, where each sample only consists of image $X^{\rm{2D,T}}$ and point cloud $X^{\rm{3D,T}}$. The images and point clouds in $\mathcal{S}$ and $\mathcal{T}$ are in the same spatial sizes, \textit{i.e.} $X^{\rm{2D}} \in \mathbb{R}^{\rm{H}\times \rm{W}\times 3}$ and $X^{\rm{3D}} \in \mathbb{R}^{\rm{N}\times 3}$. Based on these definitions, we will showcase our Mx2M.

\subsection{Network Architecture}
\label{archi}

The architecture of the Mx2M is shown in Fig.\ref{fig:arch}. For a fair comparison with previous methods \citep{jaritz2020xmuda,liu2021adversarial,peng2021sparse}, we also use the same backbone to extract features: a SparseConvNet \citep{graham20183d} for the 3D network and a modified version of U-Net \citep{ronneberger2015u} with ResNet-34 \citep{he2016deep} pre-trained on ImageNet \citep{deng2009imagenet} for the 2D one. Their output features, $H^{\rm{2D}}$ and $H^{\rm{3D}}$, have the same length N equaling the number of 3D points, where $H^{\rm{2D}}$ is gained by projecting the points into the image and sampling the 2D features at corresponding pixels. $H^{\rm{2D}}$ and $H^{\rm{3D}}$ are then sent into two groups of the same three heads, each group of which is for one modality. During these heads, the ones that predict masked 2D/3D contents $M^{\rm{2D \rightarrow 3D}}$ and $M^{\rm{3D \rightarrow 2D}}$ belong to xMRP. We will introduce them and the proceeding of masking inputs in Sec.\ref{xMRP}. Besides them, the other heads all participate in feature matching. The heads that predict final segmentation results $P^{\rm{2D}}$ and $P^{\rm{3D}}$ are our DxMFs (detailed in Sec.\ref{DxMF}). The heads that mimick the outputs from cross-modality are the linear layers inherited from xMUDA \citep{jaritz2020xmuda}, where the outputs are $P^{\rm{2D \rightarrow 3D}}$ and $P^{\rm{3D \rightarrow 2D}}$.

\begin{figure*}[t]
  \centering
  \includegraphics[width=0.97\linewidth]{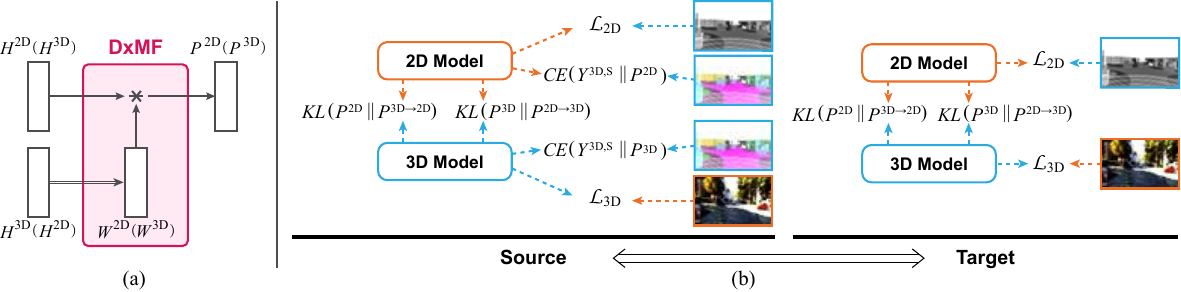}
  \caption{\label{fig:flow}(a) The details of our DxMF. The thick arrow corresponds to the modality in which the features generate dynamic weights. (b) The supervision of the source and the target domain. Mx2M introduces the cross-modal self-supervision on the target domain, which non-exists in previous methods.}
\end{figure*}

As for the information flow, we illustrate it in Fig.\ref{fig:flow}(b). The whole network is alternately trained on the source and the target domain. When the models are trained on the source domain, all six heads work. The heads for xMRP are respectively self-supervised by the origin image/point. The two DxMF heads that predict the segmentation results are both supervised by $Y^{\rm{3D,S}}$. The two mimicking heads are internally supervised by the outputs from the cross-modal DxMF heads (\textit{e.g.} $P^{\rm{3D \rightarrow 2D}}$ supervised by $P^{\rm{2D}}$). When the models are trained on the target domain, the DxMFs heads cannot be supervised because of lacking annotations. The other heads normally work as above. The loss functions of segmentation and mimicking heads are the same as previous methods \cite{jaritz2020xmuda,peng2021sparse,liu2021adversarial} for convenience, where the positions are like in Fig.\ref{fig:flow}(b). The $CE(\cdot)$ and $KL(\cdot)$ are loss functions of cross-entropy and KL divergence, respectively.

\subsection{xMRP}
\label{xMRP}
The core solution of the Mx2M, xMRP, removes a portion of the data in one modality and learns to predict the full content in the other one, which is related but different from the core solution in masked single-modality modeling. As the name implies, this procedure is divided into two steps. For the step of removal, we \textit{randomly} select some patches of the image/points and mask them inspired by the way in MAE \citep{he2021masked}. Considering that 3D points are hard to mask by patches, we first project them into the image. We use two hyper-parameters to control the masking proceeding: the $p$ indicating the size of each patch, and the $mr$ representing the masking ratio of the whole image/points (\textit{i.e.} masking $mr$ of all patches). The $mr$ cannot be as high as that in \citep{bao2021beit,he2021masked} because the CNN backbone in our method cannot perform well if the shape of objects is highly destroyed \citep{geirhos2020beyond}. Besides, due to our segmentation task, the inputs cannot always be masked and at least one modality is complete to guarantee the existence of full semantics. Thus we use another two hyper-parameters to define the ratio when masking each modality: $m_{\rm{2D}}$ meaning the ratio when masking 2D and $m_{\rm{3D}}$ indicating when masking 3D (\textit{i.e.} masking images at times of $m_{\rm{2D}}$, masking points on times of $m_{\rm{3D}}$, and no masking when (1-$m_{\rm{2D}}$-$m_{\rm{3D}}$)). We can control the inputs by ($p, mr, m_{\rm{2D}}, m_{\rm{3D}}$) to make the model adapt to different DA scenarios. As is shown in Fig.\ref{fig:arch}, $X^{\rm{2D}}$ and $X^{\rm{3D}}$ processed by these hyper-parameters (denoted as the new $X^{\rm{2D}}$ and $X^{\rm{3D}}$) are sent into the networks as inputs.

The next step is the cross-modal prediction that provides self-supervision. Inspired by the conclusion in \citep{wang2021revisiting} about the good effect of MLP on unsupervised tasks, we use the same MLP heads with middle channels of 4096 for both 2D and 3D to generate the results $M^{\rm{2D}\rightarrow{3D}}$ and $M^{\rm{3D}\rightarrow{2D}}$ for 3D and 2D, respectively. Motivated by \citep{he2021masked}, the losses are correspondingly calculated as follows:\begin{scriptsize}\begin{equation}\label{loss_sum}
\mathcal{L}_{2D}=L_2(\widehat{X}^{\rm{3D}}||M^{\rm{2D}\rightarrow{3D}}),\ \ {\rm and} \ \mathcal{L}_{3D}=L_2(\widehat{X}^{\rm{2D}}||M^{\rm{3D}\rightarrow{2D}}) .\end{equation}\end{scriptsize} The $\widehat{X}^{\rm{3D}}$ means the original 3D point clouds. The $\widehat{X}^{\rm{2D}}$ indicates the sampled pixels when $\widehat{X}^{\rm{3D}}$ projects into the original image. $L_2(\cdot)$ signs the mean squared error. It is noteworthy that we predict the full contents rather than the removed ones in masked single-modality modeling. The model can learn more 2D-3D correspondences from non-masked parts because the masked modality is different from the predicted one, which is not available in methods of masked single-modality modeling. 

Herein we finish the core proceeding of our Mx2M. The ($p, mr, m_{\rm{2D}}, m_{\rm{3D}}$) are set as (16, 0.15, 0.2, 0.2), (4, 0.3, 0.1, 0.3), and (4, 0.25, 0.3, 0.1) for scenarios of USA/Singapore, Day/Night, and A2D2/SemanticKITTI, respectively. The experiments for USA/Singapore are reported in Sec.\ref{sec:abla} and the other two are in Appendix.A. Our network can learn sufficient 2D-3D correspondences on different DA scenarios in this way, which fixes the lacking of supervision and then reduces the domain gap.

\subsection{DxMF}
\label{DxMF}

The whole network can learn more complementarity between modalities by feature matching, so it is still important for our Mx2M. Inspired by SOLO V2 \citep{wang2020solov2} which gains great progress compared with SOLO \citep{wang2020solo} via kernel-feature correspondences by locations, our DxMF constructs cross-modal kernel-feature correspondences for feature matching. The pipeline is shown in Fig.\ref{fig:flow}(a). Compared with simple final linear layers in xMUDA \citep{jaritz2020xmuda}, we use dynamic filters to segment the results. We make the procedure of segmenting the 2D results as an example to illustrate our DxMF and so do on 3D. The kernel weights $W^{\rm{2D}} \in \mathbb{R}^{\rm{N}\times F^{\rm{2D}}\times \rm{C}}$ of the filter for 2D segmentation are generated from 3D features $H^{\rm{3D}}$ by a linear layer (similarly, $W^{\rm{3D}} \in \mathbb{R}^{\rm{N}\times F^{\rm{3D}}\times \rm{C}}$ from $H^{\rm{2D}}$). As the 2D features $H^{\rm{2D}}$ have a spatial size of ($\rm{N},F^{\rm{2D}}$), the result of one point is got:\begin{equation}\label{seg}
P^{\rm{2D}}_i=W^{\rm{2D}}_i \ast H^{\rm{2D}}_i,\ \ {\rm where} \ \ i \in \rm{N} .\end{equation}

The $\ast$ indicates the dynamic convolution. We can get the segmentation results $P^{\rm{2D}}$ after all the $P^{\rm{2D}}_i$ joined together. As we dynamically construct the 2D-3D correspondences for feature matching, by which the model learns more suitable complementarity compared with the ways in previous methods \citep{jaritz2020xmuda,peng2021sparse}. We provide experiments on this comparison and ones on the scheme of the dynamic feature matching about other heads, where the results are shown in Sec.\ref{sec:abla}.

\begin{table*}[t]
    \centering
    \scriptsize
    \begin{subtable}[t]{0.33\linewidth}
    \centering
        \begin{tabular}{ccccc}
            \Xhline{1pt}
            $p$&4&8&16&32\\
            \hline
            2D&60.0&60.3&\textbf{60.4}&60.1\\
            3D&53.4&53.6&\textbf{53.8}&53.2\\
            \Xhline{1pt}
\end{tabular}
        \caption{$mr$, $m_{\rm{2D}}$, and $m_{\rm{3D}}$ are fixed.}
    \end{subtable}
    \begin{subtable}[t]{0.33\linewidth}
    \centering
    \begin{tabular}{ccccc}
            \Xhline{1pt}
            $mr$&0.15&0.20&0.25&0.10\\
            \hline
            2D&\textbf{60.4}&60.1&60.3&60.0\\
            3D&\textbf{53.8}&\textbf{53.8}&53.0&53.2\\
            \Xhline{1pt}
\end{tabular}
    \caption{$p$=16, $m_{\rm{2D}}$ and $m_{\rm{3D}}$ are fixed.}
    \end{subtable}
    \begin{subtable}[t]{0.33\linewidth}
    \centering
    \begin{tabular}{cccc}
            \Xhline{1pt}
            Head&Linear&MLP&2 MLPs\\
            \hline
            2D&61.4&\textbf{62.0}&61.5\\
            3D&56.5&\textbf{57.6}&57.4\\
            \Xhline{1pt}
\end{tabular}
    \caption{($p$, $mr$, $m_{\rm{2D}}$, $m_{\rm{3D}}$)=(16, 0.15, 0.2, 0.2).}
    \end{subtable}
    \caption{\label{tab:ablation1}Ablation studies for $p$, $mr$, and different heads for prediction correspondingly in (a), (b) and (c).}
\end{table*}
\begin{table*}[t]
\scriptsize
    \centering
\begin{tabular}{cccccccc}
            \Xhline{1pt}
             \diagbox{$m_{\rm{3D}}$}{$m_{\rm{2D}}$}&0.1&0.2&0.3&0.4&0.5&0.6&0.7\\
            \hline
            0.1&(60.4, 53.8)&(60.9, 54.1)&(61.2, 52.4)&(\textbf{61.5}, 52.1)&(60.5, 52.3)&(59.8, 51.6)&(59.7, 51.1)\\
            0.2&(60.5, 55.1)&(61.4, 56.5)&(59.4, 54.3)&(59.0, 53.9)&(58.9, 54.1)&(58.2, 52.9)&-\\
            0.3&(60.2, 54.2)&(60.0, \textbf{57.6})&(59.0, 52.7)&(58.5, 52.6)&(57.7, 51.8)&-&-\\
            0.4&(59.5, 53.5)&(58.6, 54.1)&(57.9, 52.8)&(56.7, 51.7)&-&-&-\\
            0.5&(58.6, 52.0)&(57.3, 51.9)&(57.2, 50.6)&-&-&-&-\\
            0.6&(58.0, 51.2)&(57.5, 51.0)&-&-&-&-&-\\
            0.7&(57.4, 50.1)&-&-&-&-&-&-\\
            \Xhline{1pt}
\end{tabular}
\caption{\label{tab:ablationform2d3d}Ablation for $m_{\rm{2D}}$ and $m_{\rm{3D}}$ with $p$=16 and $mr$=0.15. (mIoU 2D, mIoU 3D) correspondingly denote results for 2D and 3D networks, which are balance when $m_{\rm{2D}}$=0.2 and $m_{\rm{3D}}$=0.2.}
\end{table*}

\section{Experiments}
\label{exp}
\subsection{Implementation Details}
\label{sec:impl}
\paragraph{Datasets.}We follow three real-to-real adaptation scenarios in xMUDA \citep{jaritz2020xmuda} to implement our method, the settings of which include country-to-country, day-to-night, and dataset-to-dataset. The gaps between them raise. Three autonomous driving datasets are chosen, including nuScenes \citep{caesar2020nuscenes}, A2D2 \citep{A2D22019AEV}, and SemanticKITTI \citep{behley2019semantickitti}, where LiDAR and camera are synchronized and calibrated. In this way, we can compute the projection between a 3D point and the corresponding 2D pixel. We only utilize the 3D annotations for segmentation. In nuScenes, a point falling into a 3D bounding box is assigned the label corresponding to the object, as the dataset only contains labels for the 3D box rather than the segmentation. The nuScenes is leveraged to generate splits Day/Night and USA/Singapore, which correspond to day-to-night and country-to-country adaptation. The other two datasets are used for A2D2/SemanticKITTI ( \textit{i.e.} dataset-to-dataset adaptation), where the classes are modified as 10 according to the alignments in \citep{jaritz2020xmuda}.

\paragraph{Metrics.}Like other segmentation works, the mean intersection over union (mIoU) is adopted as the metric for evaluating the performance of the models (both 2D and 3D) for all datasets. In addition, we follow the new mIoU calculating way in \citep{jaritz2020xmuda}, which jointly considers both modalities and is obtained by taking the mean of the predicted 2D and 3D probabilities after softmax (denoted as 'Avg mIoU').

\paragraph{Inputs \& Labels.}For easily conducting masked modeling, we resize images into the sizes that could be divisible by $p$. The images in nuScenes (\textit{i.e.} Day/Night and USA/Singapore) are resized as 400$\times$224, whereas the ones in A2D2 and SemanticKITTI are reshaped as 480$\times$304. All images are normalized and then become the inputs/labels of the 2D/3D network. As for points, a voxel size of 5cm is adopted for the 3D network, which is small enough and ensures that only one 3D point lies in a voxel. The coordinates of these voxels are adopted as the labels for the 2D network.    

\paragraph{Training.}We use the PyTorch 1.7.1 framework on an NVIDIA Tesla V100 GPU card with 32GB RAM under CUDA 11.0 and cuDNN 8.0.5. For nuScenes, the mini-batch Adam \citep{kingma2015adam} is configured as the batch size of 8, $\beta_{1}$ of 0.9, and $\beta_{2}$ of 0.999. All models are trained for 100k iterations with the initial learning rate of 1e-3, which is then divided by 10 at the 80k and again at the 90k iteration. For the A2D2/SemanticKITTI, the batch size is set as 4, while related models are trained for 200k and so do on other configurations, which is caused by the limited memory. The models with '+PL' share the above proceeding, where segmentation heads are extra supervised with pseudo labels for the \textit{target} dataset. As for these pseudo labels, we strictly follow the ways in \citep{jaritz2020xmuda} to prevent manual supervision, \textit{i.e.} using the last checkpoints of models without PL to generate them offline.

\subsection{Ablation Studies}
\label{sec:abla}
To define the effectiveness of each component, we conduct ablation studies on them, respectively. As xMUDA \citep{jaritz2020xmuda} is the first method of cross-modal DA in 3D segmentation and is the baseline of all related methods \citep{liu2021adversarial,peng2021sparse}, we continue this habit and choose xMUDA as our baseline. By default, all results are reported based on the USA/Singapore scenario. For a fair comparison, we train models with each setting for 100k iterations with a batch size of 8. We also provide experiments on other scenarios, which are reported in Sec.6.

\begin{table*}[t]
    \centering
    \scriptsize
    \begin{subtable}[t]{0.33\linewidth}
    \centering
    \begin{tabular}{ccc}
            \Xhline{1pt}
            Strategy&2D&3D\\
            \hline
            xMRP&\textbf{62.0}&\textbf{57.6}\\
            2D+3D&59.4&53.3\\
            only 3D&58.9&52.6\\
            only 2D&57.9&51.9\\
            \Xhline{1pt}
    \end{tabular}
    \caption{Strategies of removal-prediction.}
    \end{subtable}
    \begin{subtable}[t]{0.33\linewidth}
    \centering
    \begin{tabular}{ccc}
            \Xhline{1pt}
            Setting&2D&3D\\
            \hline
            -&62.0&57.6\\
            +DxMF&\textbf{64.1}&\textbf{64.2}\\
            +DsCML \citep{peng2021sparse}&58.0&50.6\\
            DsCML$^{\dag}$ \citep{peng2021sparse}&57.8&50.2\\
            DsCML \citep{peng2021sparse}&61.3&53.3\\
            \Xhline{1pt}
    \end{tabular}
    \caption{Settings of cross-modal feature matching.} 
    \end{subtable}
    \begin{subtable}[t]{0.33\linewidth}
    \centering
    \begin{tabular}{ccc}
            \Xhline{1pt}
            Setting&2D&3D\\
            \hline
            DxMF on Prediction&\textbf{64.1}&\textbf{64.2}\\
            DxMF on Prediction (w/o xMRP)&61.1&53.9\\
            DxMF on Mimicking&59.4&52.4\\
            DxMF on xMRP&55.4&50.8\\
            \Xhline{1pt}
    \end{tabular}
    \caption{DxMF on three output heads.}
    \end{subtable}
    \caption{\label{tab:ablation2}Ablation for removal-prediction and DxMF in (a), (b) and (c). Three heads in (c) are mentioned in Sec.\ref{archi}.}
\end{table*}

\begin{table*}[t]
  \scriptsize
  \centering
  \begin{tabular}{clccccccccc}
    \Xhline{1pt}
     \multirow{2}*{Modality} & \multirow{2}*{Method} & \multicolumn{3}{c}{USA/Singapore} & \multicolumn{3}{c}{Day/Night} & \multicolumn{3}{c}{A2D2/SemanticKITTI} \\ 
    \cline{3-5}\cline{6-8}\cline{9-11}
    &&2D&3D&Avg&2D&3D&Avg&2D&3D&Avg\\
    \hline
    &Backbones(source only)&53.4&46.5&61.3&42.2&41.2&47.8&36.0&36.6&41.8\\
    &Backbones(on target)&66.4&63.8&71.6&48.6&47.1&55.2&58.3&71.0&73.7\\
    \hline
    \multirow{7}*{Uni-modal}&MinEnt \citep{vu2019advent}&53.4&47.0&59.7&44.9&43.5&51.3&38.8&38.0&42.7\\
    &Deep logCORAL \citep{morerio2018minimal}&52.6&47.1&59.1&41.4&42.8&51.8&35.8&39.3&40.3\\
    &PL \citep{li2019bidirectional}&55.5&51.8&61.5&43.7&45.1&48.6&37.4&44.8&47.7\\
    &FCNs in the Wild \citep{hoffman2016fcns}&53.7&46.8&61.0&42.6&42.3&47.9&37.1&43.5&43.6\\
    &CyCADA \citep{hoffman2018cycada}&54.9&48.7&61.4&45.7&45.2&49.7&38.2&43.9&43.9\\
    &AdaptSegNet \citep{tsai2018learning}&56.3&47.7&61.8&45.3&44.6&49.6&38.8&44.3&44.2\\
    &CLAN \citep{luo2019taking}&57.8&51.2&62.5&45.6&43.7&49.2&39.2&44.7&44.5\\
    \hline
    \multirow{9}*{Cross-modal}&xMUDA \citep{jaritz2020xmuda}&59.3&52.0&62.7&46.2&44.2&50.0&36.8&43.3&42.9\\
    &xMUDA+PL \citep{jaritz2020xmuda}&61.1&54.1&63.2&47.1&46.7&50.8&43.7&48.5&49.1\\
    &AUDA \citep{liu2021adversarial}&59.8&52.0&63.1&49.0&47.6&54.2&43.0&43.6&46.8\\
    &AUDA+PL \citep{liu2021adversarial}&61.9&54.8&65.6&50.3&49.7&52.6&46.8&48.1&50.6\\
    &DsCML \citep{peng2021sparse}&61.3&53.3&63.6&48.0&45.7&51.0&39.6&45.1&44.5\\
    &DsCML+CMAL \citep{peng2021sparse}&63.4&55.6&64.8&49.5&48.2&52.7&46.3&50.7&51.0\\
    &DsCML+CMAL+PL \citep{peng2021sparse}&63.9&56.3&65.1&50.1&48.7&53.0&46.8&51.8&\textbf{52.4}\\
    \cline{2-11}
    &Mx2M&64.1&64.2&64.2&49.7&49.9&49.8&44.6&48.2&47.1\\
    &Mx2M+PL&\textbf{67.4}&\textbf{67.5}&\textbf{67.4}&\textbf{52.4}&\textbf{56.3}&\textbf{54.6}&\textbf{48.6}&\textbf{53.0}&51.3\\
    \Xhline{1pt}
  \end{tabular}
  \caption{\label{tab:compare}Comparison results with both uni-modal and multi-modal adaptation methods for 3D semantic segmentation. Our Mx2M achieves state-of-the-art performance on most metrics.}
\end{table*}

\subsubsection{Ablation on xMRP}
 As mentioned in Sec.\ref{xMRP}, in xMRP, we use four hyper-parameters ($p, mr, m_{\rm{2D}}, m_{\rm{3D}}$) to control the proceeding of masking inputs and two heads of MLP to predict the cross-modality. To validate the effectiveness of the masked cross-modality modeling strategy, we insert simple xMRPs into xMUDA. The (4, 0.15, 0.1, 0.1) are selected as the start point because of the low mask ratio and the low masking 2D/3D ratio, which are suitable for the task of segmentation. As for heads, we start from the simplest \textit{linear layers}. The mIoU for (2D, 3D) in this setting are (60.0, 53.4), which are better than the segmentation results of (59.3, 52.0) in xMUDA. The good results demonstrate the significance of masked cross-modality modeling. We next explore the effectiveness of detailed settings.
 
\paragraph{Ablation on Hyper-parameters.}
To determine the suitable input settings for the current scenario, we conduct ablation studies on ($p, mr, m_{\rm{2D}}, m_{\rm{3D}}$), respectively. We start from (4, 0.15, 0.1, 0.1) and first confirm $p$ with fixed other numbers, where the mIoU of 2D and 3D are shown in Tab.\ref{tab:ablation1}(a). The networks gain the best metrics at $p=16$. The next job is to define $mr$, the results of which are illustrated in Tab.\ref{tab:ablation1}(b). Both metrics decrease with the raising of $mr$, but when $mr=0.10$ so do results. Hence the models have the best results when $mr=0.15$. Finally, we determine the $m_{\rm{2D}}$ and $m_{\rm{3D}}$. As mentioned in Sec.\ref{xMRP}, $(1-m_{\rm{2D}}$-$m_{\rm{3D}})>0$ because of keeping the full semantics. We design plenty of combinations for these two hyper-parameters, where the details are shown in Tab.\ref{tab:ablationform2d3d}. The metrics are not good when $m_{\rm{2D}}$ and $m_{\rm{3D}}$ are too large, which matches the fact that our CNN backbones cannot integrate a high mask ratio like \citep{he2021masked}. We get results of (61.4, 56.5) with suitable $m_{\rm{2D}}=0.2$ and $m_{\rm{3D}}=0.2$, and then appropriate hyper-parameters (16, 0.15, 0.2, 0.2) for the scenario.

\begin{figure*}[t]
  \centering
  \includegraphics[width=0.9\linewidth]{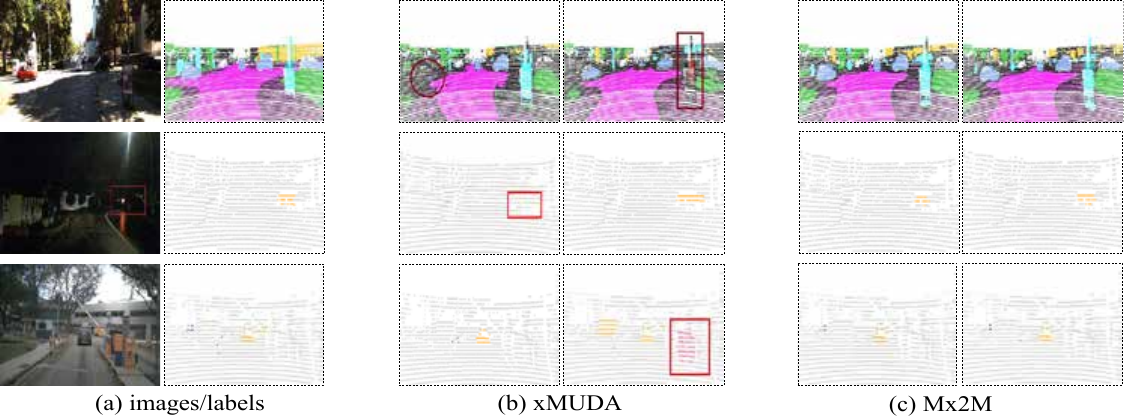}
  \caption{\label{fig:visual} Comparison of visual results on three scenarios. The images/points from the top row to the bottom come from the A2D2/SemanticKITTI, Day/Night, and USA/Singapore scenarios, respectively. With the joining of masked cross-modality modeling, errors caused by the domain gap are reduced. Best view it on screen.}
\end{figure*}

\paragraph{Ablation for Removal and Prediction.}
We obtain the results of (61.4, 56.5) with the simple linear layer. According to the conclusion in \citep{wang2021revisiting}, the network performs well when having an MLP layer. Therefore we compare the schemes of linear layer, a single MLP with mid channels of 4096, and two same MLPs with the 4096 mid channels. They are used to predict both modalities, where the results are shown in Tab.\ref{tab:ablation1}(c). A single MLP also does for our DA task. Besides, some other removal-prediction strategies are also attempted besides the cross-modal one. We illustrate the segmentation metrics in Tab.\ref{tab:ablation2}(a). We have tried respectively removing and predicting the content in 
single-modality (denoted as '2D+3D'), only in 3D point clouds, and only in 2D images. Here only removed portions are set as labels. We can see '2D+3D' has similar results as xMUDA \citep{jaritz2020xmuda}, because only rare patches work and bring about seldom information in this scheme. Similarly, the cross-modal scheme performs well thanks to 2D-3D correspondences from all contents, which is beneficial to this task \citep{peng2021sparse,li2019bidirectional}. Finally, we gain the (+2.0, +4.2) increase with our xMRP for 2D and 3D performance, respectively.

\subsubsection{Ablation on DxMF}
All above experiments are based on the same way of feature matching as xMUDA \citep{jaritz2020xmuda}, where the segmentation results are got based on two linear layers. We also conduct experiments on our DxMF, which achieves cross-modal feature matching and then the segmentation by dynamically constructing kernel-feature correspondences. The comparison is shown in the first two rows of Tab.\ref{tab:ablation2}(b), our DxMF performs the better 2D-3D complementarity and especially increases the 3D performance. We also try to combine the means of sparse-to-dense cross-modal feature matching, DsCML \citep{peng2021sparse}, with the masked cross-modality modeling, where the metrics are illustrated in the last three rows in Tab.\ref{tab:ablation2}(c). The results with '$\dag$' or not denote that they are from the implementation of the official source code or from the paper. As our experiments are based on the official source code, we still gain the increase with the join of xMRP. In all, the good metrics prove the effectiveness of our DxMF.

We also validate the results for only using DxMF, which is reported in the first two rows of Tab.\ref{tab:ablation2}(c). Besides, our Mx2M has three output heads for each modality according to Sec.\ref{archi}. We also conduct experiments on DxMF on them besides the above experiments on prediction heads. Like adding DxMF to prediction heads, we add DxMF to other ones on both modalities. The results are reported in Tab.\ref{tab:ablation2}(c). Both mimicking heads and ones for xMRP do not match the DxMF. We may infer that the former is not involved in segmentation and the latter is like respective single-modal prediction in Tab.\ref{tab:ablation2}(a). Both situations are not suitable for our DxMF. After all experiments, our Mx2M outperforms the Baseline xMUDA (+4.8, +12.2) for 2D and 3D in total, which shows that the Mx2M does work.

\subsection{Limitations}
\label{sec:limitations}
Considering previous works \citep{liu2021adversarial,peng2021sparse} attempt to introduce the adversarial learning (AL) into the DA in 3D semantic segmentation, we also add the extra heads for AL in both 2D and 3D. We use the simple AL in AUDA \citep{liu2021adversarial} and the CMAL in DsCML \citep{peng2021sparse}. The results for 2D and 3D are not ideal, which are correspondingly (56.26, 51.76) and (49.75, 41.94) for AL in AUDA and CMAL. Compared with the metrics of (64.1, 64.2) in the scheme without AL, they decrease so much. We think it is the limitation in our Mx2M that our method does not match AL.
\subsection{Comparison with The State-of-the-art}
We evaluate our Mx2M on the above three real-to-real DA scenarios and compare the results with some methods. First, we train the backbones on source only and on target only (except on the Day/Night, where the batches of 50\%/50\% Day/Night are used to prevent overfitting). The two results can be seen as the upper and the lower limit of the DA effectiveness. Next, some representative uni-modal DA methods are compared. These uni-modal methods are correspondingly evaluated on U-Net with ResNet-34 for 2D and SparseConvNet for 3D, which are the same as our backbones. We use the results from \citep{peng2021sparse} for convenience. Finally, We also compare our method with some cross-modal methods, including xMUDA \citep{jaritz2020xmuda}, AUDA \citep{liu2021adversarial}, and DsCML \citep{peng2021sparse}. These cross-modal methods and our Mx2M are also trained with the data with pseudo labels on the target domain, where the proceeding can be seen in Sec.\ref{sec:impl}. 

All comparison results for 3D segmentation are reported in Tab.\ref{tab:compare}. We can see that the Mx2M gains the (2D mIoU, 3D mIoU) on average of (+5.4, +7.6) compared with the baseline xMUDA, which proves the DA performance of our method. Specifically, for the USA/Singapore scenario, the bare Mx2M even surpasses xMUDA with PL. In Day/Night, though the metric without PL looks normal, the result with PL shows a surprising increase that is close to the upper limit. As for the A2D2/SemanticKITTI, the Mx2M outperforms all methods on 2D and 3D metrics with a 0.9 less Avg mIoU compared to the DsCML. In total, our Mx2M gains state-of-the-art performance on most metrics. We also provide some visual results, which are shown in Fig.\ref{fig:visual}. More visual results can be seen in Sec.7.

\section{Conclusion}

In this paper, we propose a method named Mx2M for domain adaptation in 3D semantic segmentation, which utilizes masked cross-modality modeling to solve the problem of lacking supervision on the target domain and then reduce the large gap. The Mx2M includes two components. The core solution xMRP makes the Mx2M adapts to various scenarios and provides cross-modal self-supervision. A new way of cross-modal feature matching named DxMF ensures that the whole method exploits more suitable 2D-3D complementarity and then segments results. We achieve state-of-the-art performance on three DA scenarios for 3D segmentation, including USA/Singapore, Day/Night, and A2D2/SemanticKITTI. Specifically, the Mx2M with pseudo labels achieves the (2D mIoU, 3D mIoU, Avg mIoU) of (67.4, 67.5, 67.4), (52.4, 56.3, 54.6), and (48.6, 53.0, 51.3) for the three scenarios. All the above results demonstrate the effectiveness of our method.

\section{Ablation Studies on Scenarios of Day/Night and A2D2/SemanticKITTI}
We also conduct ablation studies on the scenarios of Day/Night and A2D2/SemanticKITTI. Similarly, the xMUDA \citep{jaritz2020xmuda} is also selected as the backbone. We train the models with each Day/Night setting for 100k iterations with a batch size of 8 and with each A2D2/SemanticKITTI setting for 200k iterations with a batch size of 4 because of limited resources. The proceedings are basically as those in USA/Singapore and as follows.
\subsection{Ablations on Day/Night}
\label{ablationonday}
We first validate the effectiveness of the masked cross-modality modeling strategy. The four hyper-parameters ($p, mr, m_{\rm{2D}}, m_{\rm{3D}}$) are set as (4, 0.15, 0.1, 0.1), which is the same as what we do for the USA/Singapore and for the same reason in Sec.4.2.1 of the body. As for the heads for predicting segmentation, we start from the simplest \textit{linear layers}. The mIoU results for (2D, 3D) are (47.3, 45.9), which are better than the segmentation indexes of (46.2, 44.2) in our baseline xMUDA. The good results demonstrate the significance of the strategy for masked cross-modality modeling. We next explore the effectiveness of detailed settings.

To determine the suitable input settings for the current scenario, we conduct ablation studies on ($p, mr, m_{\rm{2D}}, m_{\rm{3D}}$), respectively. We start from (4, 0.15, 0.1, 0.1) and first confirm $p$ with fixed other numbers, where the mIoU of 2D and 3D are shown in Tab.\ref{tab:ablation12}(a). The start point is the most suitable one for this scenario, \textit{i.e.} $p=4$. The next job is to define $mr$, the results of which are illustrated in Tab.\ref{tab:ablation12}(b). The models have the best results when $mr=0.3$. Finally, we determine the $m_{\rm{2D}}$ and $m_{\rm{3D}}$. We design plenty of combinations for these two hyper-parameters, where the details are shown in Tab.\ref{tab:ablationform2d3d12}. We get results of (48.9, 48.5) with suitable numbers, where $m_{\rm{2D}}=0.1$ and $m_{\rm{3D}}=0.3$, and then appropriate hyper-parameters (4, 0.3, 0.1, 0.3) for the scenario.

We obtain the results of (48.9, 48.5) with the simple linear layer. According to the conclusion in \citep{wang2021revisiting}, the network performs well when having an MLP layer. Therefore we compare the schemes of linear layer, a single MLP with mid channels of 4096, and two same MLPs with the 4096 mid channels. They are used to predict both modalities, where the results are shown in Tab.\ref{tab:ablation12}(c). A single MLP also does for our DA task. Finally, the DxMF is added to the Mx2M, the results of which are illustrated in Tab.\ref{tab:ablation12}(d). With the join of the DxMF, we gain the mIoU of (49.7, 49.9) for the scenario of Day/Night.
\subsection{Ablations on A2D2/SemanticKITTI}
The proceeding of ablation studies on A2D2/SemanticKITTI is the same as what in Day/Night. First, the effectiveness of the masked cross-modality modeling strategy is validated. Our results are (37.9, 44.3) \textit{v.s.} (36.8, 43.3) of the ones in xMUDA \citep{jaritz2020xmuda}. Next, we determine four hyper-parameters ($p, mr, m_{\rm{2D}}, m_{\rm{3D}}$), where the procedure is the same as the above. We report them in Tab.\ref{tab:ablation22}(a), Tab.\ref{tab:ablation22}(b), and Tab.\ref{tab:ablationform2d3d2}, respectively. They are set as (4, 0.25, 0.3, 0.1) for the A2D2/SemanticKITTI scenario. We then define the prediction heads with results (41.5, 46.2), which is shown in Tab.\ref{tab:ablation22}(c). The MLP works again. Finally, we validate the effectiveness of the DxMF, the results of which are illustrated in Tab.\ref{tab:ablation22}(d). We gain the mIoU of (44.6, 48.2) for the A2D2/SemanticKITTI.

\section{More Visual Results on the Mx2M}
\begin{figure*}[h]
  \centering
  \includegraphics[width=1\linewidth]{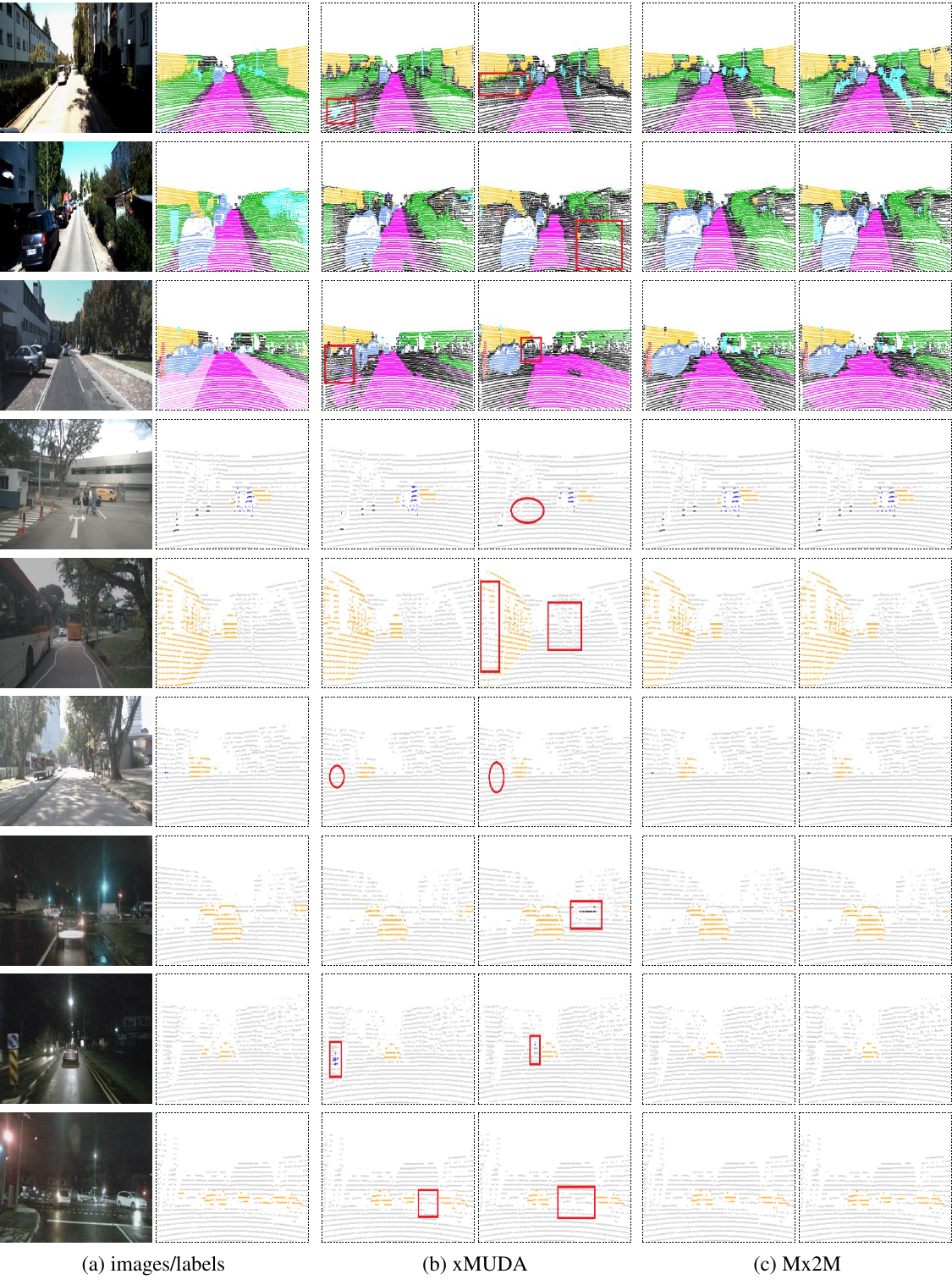}
  \caption{\label{fig:visualapp} Comparison of visual results on three scenarios. Our Mx2M achieves better domain adaptation and more precise segmentation.}
\end{figure*}

As is mentioned in Sec.4.4 of the main body, we offer more visual results in Fig.\ref{fig:visualapp}. The images/points from the top row to the bottom correspondingly come from the A2D2/SemanticKITTI, USA/Singapore, and Day/Night scenarios, where every three rows belong to the same scenario. Our Mx2M has a balanced performance in all three scenarios.
\begin{table*}[t]
    \centering
    \scriptsize
    \begin{subtable}[t]{0.2\linewidth}
    \centering
    \begin{tabular}{cccc}
            \Xhline{1pt}
            $p$&4&8&2\\
            \hline
            2D&\textbf{47.3}&47.2&46.6\\
            3D&\textbf{45.9}&45.2&45.8\\
            \Xhline{1pt}
    \end{tabular}
    \caption{\label{tab:ablationforp}$mr$, $m_{\rm{2D}}$, and $m_{\rm{3D}}$ are fixed.}
    \end{subtable}\quad
    \begin{subtable}[t]{0.3\linewidth}
    \centering
    \begin{tabular}{cccccc}
            \Xhline{1pt}
            $mr$&0.15&0.20&0.25&0.30&0.35\\
            \hline
            2D&47.3&47.2&47.6&\textbf{47.9}&47.2\\
            3D&45.9&46.1&46.2&\textbf{46.5}&46.3\\
            \Xhline{1pt}
    \end{tabular}
    \caption{\label{tab:ablationformr}$p=4$, $m_{\rm{2D}}$ and $m_{\rm{3D}}$ are fixed.}
    \end{subtable}\quad
    \begin{subtable}[t]{0.25\linewidth}
    \centering
    \begin{tabular}{cccc}
           \Xhline{1pt}
            Head&Linear&MLP&2 MLPs\\
            \hline
            2D&48.9&\textbf{49.5}&49.3\\
            3D&48.5&\textbf{49.3}&48.9\\
            \Xhline{1pt}
    \end{tabular}
    \caption{\label{tab:ablationforheads}($p$, $mr$, $m_{\rm{2D}}$, $m_{\rm{3D}}$)=(4, 0.3, 0.1, 0.3).}
    \end{subtable}\quad
    \begin{subtable}[t]{0.2\linewidth}
    \centering
    \begin{tabular}{ccc}
            \Xhline{1pt}
            Setting&2D&3D\\
            \hline
            -&49.5&49.3\\
            +DxMF&\textbf{49.7}&\textbf{49.9}\\
            \Xhline{1pt}
    \end{tabular}
    \caption{\label{tab:ablationfordaynight}Ablation for DxMF.} 
    \end{subtable}
    \caption{\label{tab:ablation12}Ablation studies for $p$, $mr$, different heads, and DxMF about Day/Night for prediction correspondingly in (a), (b), (c) and (d).}
\end{table*}

\begin{table*}[t]
\scriptsize
    \centering
\begin{tabular}{cccccccc}
            \Xhline{1pt}
             \diagbox{$m_{\rm{3D}}$}{$m_{\rm{2D}}$}&0.1&0.2&0.3&0.4&0.5&0.6&0.7\\
            \hline
            0.1&(47.9, 46.5)&(48.0, 46.6)&(48.6, 47.1)&(48.3, 47.5)&(47.8, 47.9)&(47.2, 46.9)&(47.6, 46.7)\\
            0.2&(48.5, 47.9)&(48.8, 48.0)&(48.8, 48.4)&(48.4, 46.7)&(47.7, 47.5)&(46.8, 46.6)&-\\
            0.3&(\textbf{48.9}, \textbf{48.5})&(48.6, 47.9)&(48.3, 46.7)&(47.6, 47.9)&(48.3, 45.2)&-&-\\
            0.4&(48.8, 48.1)&(47.5, 46.8)&(47.8, 45.2)&(47.9, 47.0)&-&-&-\\
            0.5&(47.1, 48.4)&(47.0, 46.5)&(48.0, 43.9)&-&-&-&-\\
            0.6&(47.2, \textbf{48.5})&(46.0, 46.7)&-&-&-&-&-\\
            0.7&(46.6, 48.3)&-&-&-&-&-&-\\
            \Xhline{1pt}
\end{tabular}
 \caption{\label{tab:ablationform2d3d12}Ablation for $m_{\rm{2D}}$ and $m_{\rm{3D}}$ with $p=4$ and $mr=0.3$ in the Day/Night. (mIoU 2D, mIoU 3D) denote results of corresponding networks, which are better when $m_{\rm{2D}}=0.1$ and $m_{\rm{3D}}=0.3$.}
\end{table*}

\begin{table*}[t]
    \centering
    \scriptsize
    \begin{subtable}[t]{0.2\linewidth}
    \centering
    \begin{tabular}{cccc}
            \Xhline{1pt}
            $p$&4&8&2\\
            \hline
            2D&\textbf{37.9}&37.1&37.4\\
            3D&\textbf{44.3}&43.6&43.2\\
            \Xhline{1pt}
    \end{tabular}
    \caption{\label{tab:ablationforp2}Ablation for $p$ with fixed $mr$, $m_{\rm{2D}}$.}
    \end{subtable} \quad  
    \begin{subtable}[t]{0.25\linewidth}
    \centering
    \begin{tabular}{ccccc}
            \Xhline{1pt}
            $mr$&0.15&0.20&0.25&0.30\\
            \hline
            2D&37.9&38.3&\textbf{39.1}&38.0\\
            3D&44.3&44.4&\textbf{45.0}&44.0\\
            \Xhline{1pt}
    \end{tabular}
    \caption{\label{tab:ablationformr2}Ablation for $mr$ with $p=4$ and fixed $m_{\rm{2D}}$ as well as $m_{\rm{3D}}$.}
    \end{subtable}\quad
    \begin{subtable}[t]{0.25\linewidth}
    \centering
    \begin{tabular}{cccc}
            \Xhline{1pt}
            Head&Linear&MLP&2 MLPs\\
            \hline
            2D&41.5&\textbf{43.6}&43.0\\
            3D&46.2&\textbf{47.8}&46.8\\
            \Xhline{1pt}
    \end{tabular}
    \caption{\label{tab:ablationforheads2}Ablation on different heads for prediction.}
    \end{subtable} \quad   
    \begin{subtable}[t]{0.2\linewidth}
    \begin{tabular}{ccc}
            \Xhline{1pt}
            Setting&2D&3D\\
            \hline
            -&43.6&47.8\\
            +DxMF&\textbf{44.6}&\textbf{48.2}\\
            \Xhline{1pt}
    \end{tabular}
    \caption{\label{tab:ablationfora2d2}Ablation for DxMF.}
    \end{subtable}
    \caption{\label{tab:ablation22}Ablation studies for $p$, $mr$, different heads, and DxMF about A2D2/SemanticKITTI for prediction correspondingly in (a), (b), (c) and (d)..}
\end{table*}

\begin{table*}[t]
\scriptsize
    \centering
\begin{tabular}{cccccccc}
            \Xhline{1pt}
             \diagbox{$m_{\rm{3D}}$}{$m_{\rm{2D}}$}&0.1&0.2&0.3&0.4&0.5&0.6&0.7\\
            \hline
            0.1&(39.1, 45.0)&(40.2, 45.5)&(\textbf{41.5}, \textbf{46.2})&(40.9, 45.5)&(40.4, 45.3)&(40.3, 44.9)&(39.2, 45.4)\\
            0.2&(35.6, 42.8)&(40.3, 44.1)&(40.8, 44.9)&(39.8, 45.1)&(39.6, 45.0)&(40.4, 40.3)&-\\
            0.3&(35.3, 45.6)&(40.8, 43.4)&(39.9, 44.4)&(39.1, 43.4)&(39.7, 41.6)&-&-\\
            0.4&(37.0, 41.8)&(39.9, 44.3)&(38.1, 43.0)&(37.4, 44.9)&-&-&-\\
            0.5&(40.3, 45.5)&(39.1, 45.2)&(39.0, 42.9)&-&-&-&-\\
            0.6&(37.1, 45.5)&(37.3, 44.4)&-&-&-&-&-\\
            0.7&(36.7, 44.8)&-&-&-&-&-&-\\
            \Xhline{1pt}
\end{tabular}
\caption{\label{tab:ablationform2d3d2}Ablation for $m_{\rm{2D}}$ and $m_{\rm{3D}}$ with $p=4$ and $mr=0.25$ in the A2D2/SemanticKITTI. (mIoU 2D, mIoU 3D) denote results of corresponding networks, which are better when $m_{\rm{2D}}=0.3$ and $m_{\rm{3D}}=0.1$.}
\end{table*}



\bibliography{aaai23}
\end{document}